\documentclass[letterpaper]{article} 
\usepackage{aaai25}  
\usepackage{times}  
\usepackage{helvet}  
\usepackage{courier}  
\usepackage[hyphens]{url}  
\usepackage{graphicx} 
\usepackage{natbib}  
\usepackage{caption} 
\usepackage{algorithm}
\usepackage{algorithmic}

\usepackage{amsthm}
\usepackage{amsmath}

\usepackage{enumitem}
\usepackage{booktabs}
\usepackage{xcolor}
\usepackage{mathtools}
\definecolor{mygrey}{cmyk}{0, 0, 0, 59}

\usepackage{multirow}
\usepackage{amssymb}
\usepackage{bm}
\usepackage{colortbl}
\usepackage{tabularx}
\usepackage{listings}

\usepackage{newfloat}
\usepackage{listings}
\DeclareCaptionStyle{ruled}{labelfont=normalfont,labelsep=colon,strut=off} 
\floatstyle{ruled}
\newfloat{listing}{tb}{lst}{}
\floatname{listing}{Listing}
\title{Concept Matching with Agent for Out-of-Distribution Detection}
\author{
    Yuxiao Lee\textsuperscript{\rm 1}, Xiaofeng Cao\textsuperscript{\rm 1}\thanks{Corresponding author.}, Jingcai Guo\textsuperscript{\rm 2}, Wei Ye\textsuperscript{\rm 3}, Qing Guo\textsuperscript{\rm 4}, Yi Chang\textsuperscript{\rm 1, 5}
}
\affiliations{
    \textsuperscript{\rm 1}School of Artificial Intelligence, Jilin University, China\\
    \textsuperscript{\rm 2}The Hong Kong Polytechnic University, \textsuperscript{\rm 3}College of Electronic and Information Engineering, Tongji University, China\\
    \textsuperscript{\rm 4}CFAR and IHPC, Agency for Science, Technology and Research (A*STAR), Singapore\\
    \textsuperscript{\rm 5}Engineering Research Center of Knowledge-Driven Human-Machine Intelligence, Ministry of Education, China
    yuxiao9922@mails.jlu.edu.cn, xiaofengcao@jlu.edu.cn,\\ jc-jingcai.guo@polyu.edu.hk, yew@tongji.edu.cn, tsingqguo@ieee.org, yichang@jlu.edu.cn
}

\usepackage{bibentry}

\begin{document}

\maketitle

\begin{abstract}
The remarkable achievements of Large Language Models (LLMs) have captivated the attention of both academia and industry, transcending their initial role in dialogue generation. To expand the usage scenarios of LLM, some works enhance the effectiveness and capabilities of the model by introducing more external information, which is called the agent paradigm. Based on this idea, we propose a new method that integrates the agent paradigm into out-of-distribution (OOD) detection task, aiming to improve its robustness and adaptability. Our proposed method, Concept Matching with Agent (CMA), employs neutral prompts as agents to augment the CLIP-based OOD detection process. These agents function as dynamic observers and communication hubs, interacting with both In-distribution (ID) labels and data inputs to form vector triangle relationships. This triangular framework offers a more nuanced approach than the traditional binary relationship, allowing for better separation and identification of ID and OOD inputs. Our extensive experimental results showcase the superior performance of CMA over both zero-shot and training-required methods in a diverse array of real-world scenarios.
\end{abstract}

%
\begin{links}
    \link{Code}{https://github.com/yuxiaoLeeMarks/CMA}
\end{links}

\section{Introduction}
\label{sec:introduction}

The emergence and development of Large Language Models (LLMs) \citep{chang2024survey,zhao2023survey,gpt3,gpt4} have significantly reshaped the landscape of Artificial Intelligence (AI), marking a pivotal breakthrough in both academic research and practical applications. These models have not only revolutionized the way we generate conversations but also demonstrated their capacity as intermediary agents with more nuanced roles, facilitating the accomplishment of myriad tasks with unprecedented efficiency and adaptability \citep{wang2024survey,xi2023rise,zeng2023agenttuning}. The Agent paradigm has been extensively applied across multiple domains and tasks, playing a profound role \citep{qian2023communicative,qian2023experiential,hong2023metagpt}. The core of this paradigm lies in introducing \textbf{external information to change the input distribution of the model}, thereby enhancing the model's performance.

\begin{figure}[h]
  \centering
  \includegraphics[width=0.33\textwidth]{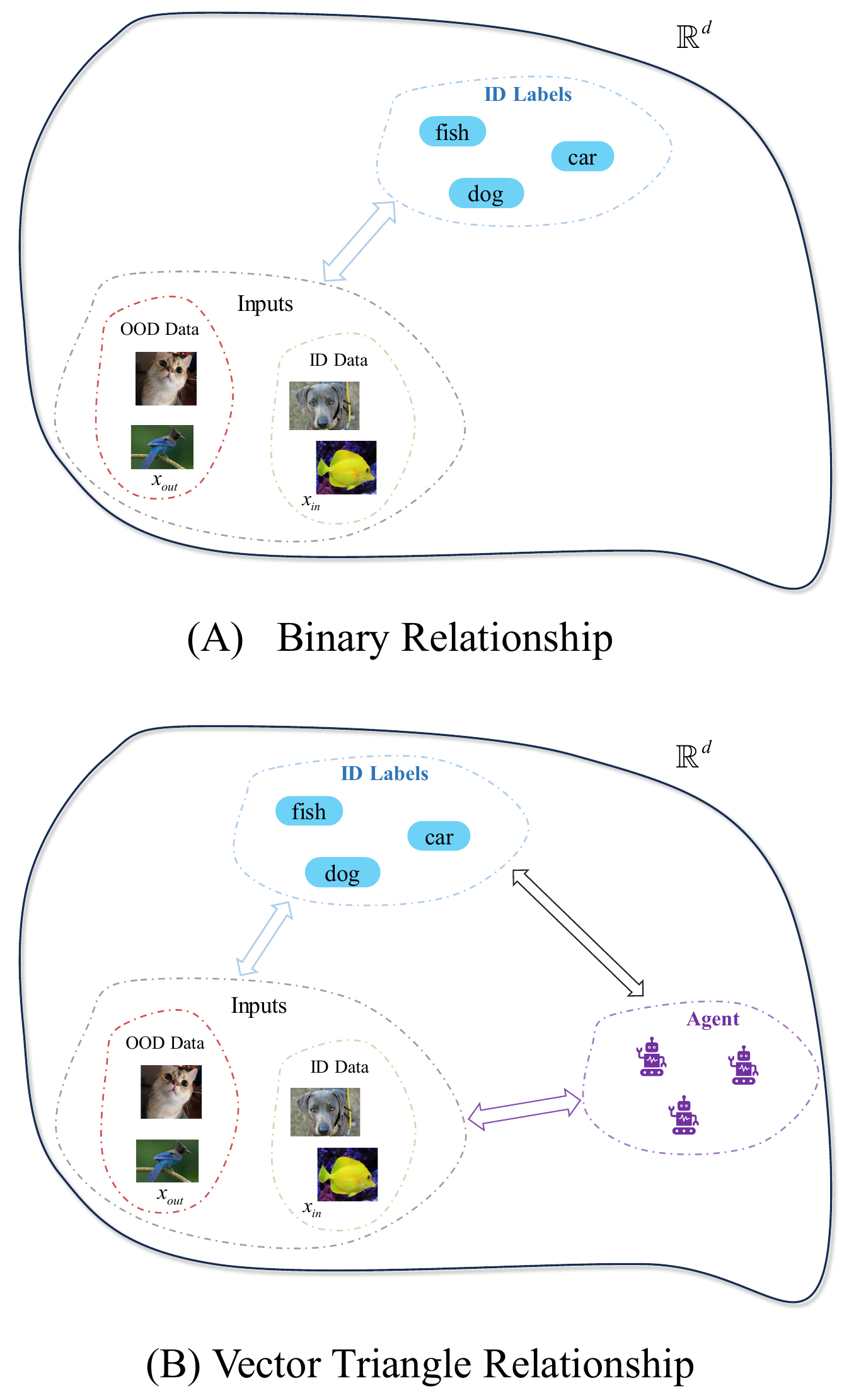}
  \caption{In OOD detection, the Vector Triangle Relationship alters the traditional Binary Relationship by introducing Agents, thereby more effectively processing and distinguishing between ID data and OOD data.}
  \label{fig:visual_example}
\end{figure}

An important and challenging task within the field of machine learning is to enhance the robustness of models across diverse scenarios. When an artificial intelligence system encounters data that significantly deviates from its training data distribution, OOD detection becomes crucial for ensuring its reliability and robustness. In the past, most OOD detection methods employed single-modal learning \citep{survey1,survey2,hendrycks2016baseline}. As CLIP \citep{CLIP} has demonstrated astonishing performance across various downstream tasks, an increasing number of CLIP-based methods for out-of-distribution (OOD) detection have emerged \citep{ramesh2022hierarchical,wang2022clip,crowson2022vqgan,wang2021actionclip,gao2024clip}. 
\paragraph{Question.} However, previous OOD detection methods, whether single-modal learning or CLIP-based approaches, typically rely on binary relationship to differentiate between in-distribution (ID) data and OOD data (Figure \ref{fig:visual_example}). These methods include solely using ID data to construct boundaries or employing a combination of ID and OOD data to demarcate their respective domains. While these methods are effective to some extent, they lack the flexibility and adaptability needed to handle the dynamic complexity of real-world data distributions.

\begin{figure*}
  \centering
  \resizebox{\textwidth}{!}{\includegraphics{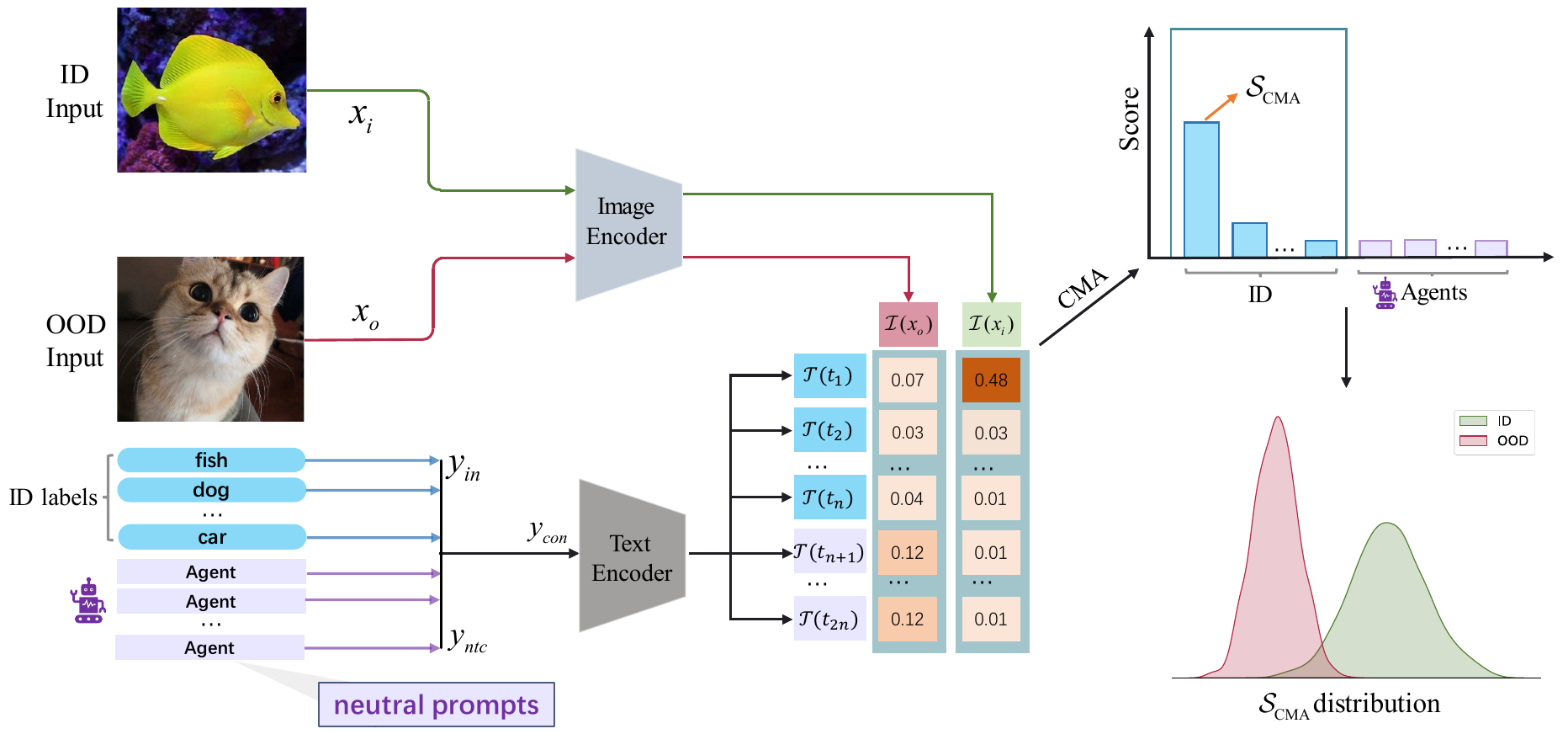}}
  \caption{\textbf{Overview of Concept Matching with Agent (CMA) framework.} The input image $x$ undergoes Image Encoder $\mathcal{I}$ to produce an image embedding. The concatenation of the ID labels $\mathcal{Y}_{in}$ and Agents $\mathcal{Y}_{ntc}$ is then subjected to Text Encoder $\mathcal{T}$ to generate a text embedding. The similarity between the image and text embeddings is computed, with a higher result indicating a greater degree of similarity (darker shading denotes higher similarity). This is followed by the CMA operation, which computes the $\mathcal{S}_{\text{CMA}}$ for each image as the ultimate discriminative metric. Further details are provided in Section \ref{sec:method}.}
  \label{fig1}
\end{figure*}


\paragraph{Motivation.} 
In the realm of  OOD  detection, the predominant methodology consists of binary segmentation between  ID  and OOD data, typically facilitated by scoring functions. The effectiveness of this prototype is closely tied to the sophisticated design of these scoring functions. Inspired by the remarkable success of the agent-based paradigm, introducing external information as agents in  CLIP-based OOD detection framework could reshape the distribution of ID and OOD inputs. Structurally, this paradigm shift from binary segmentation to a vector triangle relationship relational framework holds substantial potential for uncovering deeper insights, potentially transforming the interplay between ID labels and data inputs (Figure ~\ref{fig:visual_example}).

\paragraph{Our scheme.} In this paper, we propose the \textbf{C}oncept \textbf{M}atching with \textbf{A}gent (\textbf{CMA}) methodology, which integrates \textbf{\textit{neutral textual concept prompts in natural language}} as \textbf{\textit{Agents}} within the CLIP-based OOD detection framework. Our framework is illustrated in Figure~\ref{fig1}. These agents serve dual roles: as observers and as intermediate hubs that facilitate the interaction between ID Labels and Data inputs. By doing so, we aim to establish a vector triangle relationship among the ID labels, data inputs, and the agents themselves. In this triangular vector relationship (Figure ~\ref{fig:visual_example}), the score of OOD data is diminished due to the collision effect of the Agents, thereby widening the gap between the scores of ID data and OOD data. In other words, images closer to the ID class are less likely to be influenced by neutral text concepts, whereas images from the OOD class are more susceptible to these concepts, resulting in lower scores. The method to achieve our idea is derived from our profound insights into the Language-Vision representation (See Section \ref{sec:method}). Building on these insights, we formulate the entire Concept Matching with Agent (CMA) framework.

In summary, our proposed method, CMA, possesses three distinct advantages. (1) Our approach obviates the need for training data, enabling zero-shot OOD detection while leveraging the collision between Agent and both ID and OOD data to effectively widen the gap between the two, ensuring robust performance. This stands in stark contrast to traditional OOD detection methods, which rely on extensive external data for intricate training \citep{survey1,survey2}. (2) Our method exhibits remarkable scalability, allowing for the tailoring of specialized Agents to suit various scenarios, thereby further enhancing performance (See Section \ref{sec:discussion}). This is facilitated by the flexibility of our triangular vector relationship, which can enhance the impact of certain OOD images through specific Agents, thereby reducing their scores. (3) It is noteworthy that the CMA maintains robustness against both hard OOD inputs, encompassing both semantic hard OODs \citep{semantically_hard_ood} and spurious OODs \citep{spurious_OOD}. This makes our approach a truly practical and viable option. The contributions of our study are summarized as follows:
\begin{itemize}
      \item Drawing upon the concept of Agents in  LLMs, we propose the incorporation of agent-based observation into OOD detection. By facilitating interactions among agents,  ID  labels, and data inputs, we establish a vector triangle relationship for them. This structured shift diverges from the conventional binary frameworks, offering enhanced flexibility in application scenarios and providing a novel analytical perspective on OOD detection.
      \item We propose a novel CLIP-based OOD Detection framework. Compared to previous methods, our approach more effectively achieves the objective of OOD Detection: it widens the gap between ID and OOD, and possesses enhanced versatility and practicality.
      \item We conducted experiments on various datasets with distinct ID scenarios and demonstrated that CMA achieves superior performance across a wide range of real-world tasks. Compared to most existing OOD detection methods, CMA brings substantial improvements to the large-scale ImageNet OOD benchmark.
\end{itemize}

\section{Preliminaries}
\label{sec:perliminaries}

\paragraph{Contrastive Vision-Language Models.} In comparison to traditional CNN architectures, the ViT \citep{vit} leverages the Transformer Encoder framework \citep{transformer} to accomplish the task of image classification, realizing the possibility of utilizing language model architectures for visual tasks. This provides insights into the field of vision-language representation learning, with CLIP \citep{CLIP} being a notable representative. CLIP employs self-supervised contrastive objectives to embed images and their corresponding textual descriptions into a shared feature space, achieving alignment between the two. Structurally, CLIP, which utilizes a dual-stream architecture, comprises an image encoder $\mathcal{I}:x \to \mathbb{R}^d$ and a text encoder $\mathcal{T}:t \to \mathbb{R}^d$. After pretraining on a dataset of 400 million text-image pairs, the joint visual-language embedding of CLIP associates objects in various patterns. Due to the robust performance of CLIP, several OOD detection methods based on it have emerged. Nevertheless, the challenge of how to better utilize Language-Vision representation for OOD detection remains a difficult yet significant issue.

\paragraph{Zero-shot OOD Detection.} For traditional OOD detection frameworks \citep{hendrycks2016baseline}, a common assumption is made of a typical real-world scenario wherein classifier $f$ are trained on ID data categorized as $\mathcal{Y}_{in} = \left\{1, ..., C\right\}$ and subsequently deployed in an environment containing samples from unknown classes $y \notin \mathcal{Y}_{in}$, outside the distribution of the ID. The classifier $f$ is then tasked with determining membership to the ID. In contrast, zero-shot OOD detection \citep{MCM,glmcm,fort2021exploring}, in line with the current trend of deep learning, leverages pre-trained models on open datasets, eliminating the need for additional training of the model. The approach determines membership to the ID by calculating results through mapping the data into a common space $\mathbb{R}^d$. Compared to traditional OOD detection frameworks that necessitate training, the zero-shot OOD detection framework is notably more versatile and practical.

\paragraph{CLIP-based OOD Detection}

The CLIP model aligns image features with text features describing the image in a high-dimensional space by simultaneously training image and text encoders on a large dataset, thereby learning rich visual-language joint representations. When applied to OOD detection tasks, CLIP only requires class names and does not require training on specific ID data, allowing it to attempt to classify or determine whether an input image belongs to a known class. It is worth noting that the ID classes in CLIP-based OOD detection refer to the classes used in downstream classification tasks, which are different from the pre-trained classes in the upstream. The OOD classes are those that do not belong to any of the ID classes in the downstream tasks.

For current CLIP-based OOD detection, MCM \citep{MCM} has become a basic paradigm. Its core idea is to treat text embeddings as ``concept prototypes" and evaluate their in-distribution or out-of-distribution properties by measuring the similarity between the input image features and these concept prototypes. Specifically, given a set of ID categories $\mathcal{Y}_{in}$ with a corresponding text description for each category, we first use a pre-trained text encoder $\mathcal{T}$ to convert these text descriptions into $d$ dimensional vectors $\mathbf{c}_i = \mathcal{T}(t_{i}) \in \mathbb{R}^d$, where $i \in \{1,2,...,N \}$ and $N$ is the number of ID categories. For any input image $\mathbf{x}^{\prime}$ whose visual features are extracted by an image encoder $\mathcal{I}$ as $\mathbf{v}' = \mathcal{I}(\mathbf{x}^{\prime}) \in \mathbb{R}^d$, the MCM score is defined as the cosine similarity between the visual features and the closest concept prototype, which is scaled by an appropriate softmax to enhance the separability of ID and OOD samples:

\begin{align*}
\mathcal{S}_{\text{MCM}}(\mathbf{x}^{\prime}; \mathcal{Y}_{in}, \mathcal{T}, \mathcal{I}) = \frac{\exp(\text{sim}(\mathbf{v}', \mathbf{c}_{\hat{y}})/\tau)}{\sum_{i=1}^{N}\exp(\text{sim}(\mathbf{v}', \mathbf{c}_i)/\tau)},
\end{align*}
where $\text{sim}(\cdot, \cdot)$ denotes cosine similarity, $\hat{y} = \arg\max_i \text{sim}(\mathbf{v}', \mathbf{c}_i)$ denotes the most matching concept index, and $\tau$ is a temperature parameter used to adjust the distribution of similarity.

\section{Method}
\label{sec:method}

\subsection{Out-of-Distribution Detection}

Out-of-Distribution (OOD) detection pertains to the task of discerning whether a given input sample originates from the same distribution as the training data (in-distribution, ID) or from a different distribution (out-of-distribution, OOD). Formally, let $ X $ denote the input space, with $ P_{\text{in}} $ denoting the probability distribution of ID data and $ P_{\text{out}} $ denoting the distribution of OOD data. The objective of OOD detection is to classify an input sample $ x \in X $ as either belonging to $ P_{\text{in}} $ or $ P_{\text{out}} $. This problem can be framed as a binary classification challenge, where the model output, denoted as $ f(x; \theta) $, yields a probability score $ p $ indicating the confidence that $ x $ is in-distribution: $ p(x) = P(y=1 | x; \theta) $, with $ y=1 $ signifying that $ x $ is in-distribution. A threshold $ \tau $ is established to facilitate classification, whereby if $ p(x) > \lambda $, the sample is classified as in-distribution, and if $ p(x) \leq \lambda $, it is classified as out-of-distribution. Thus, OOD detection aims to enhance the robustness and reliability of machine learning models by effectively mitigating the risks posed by unseen or anomalous data.

\subsection{How to construct vector triangle relationships using Agents}

\begin{figure*}[t]
  \centering
    \resizebox{\textwidth}{!}{\includegraphics{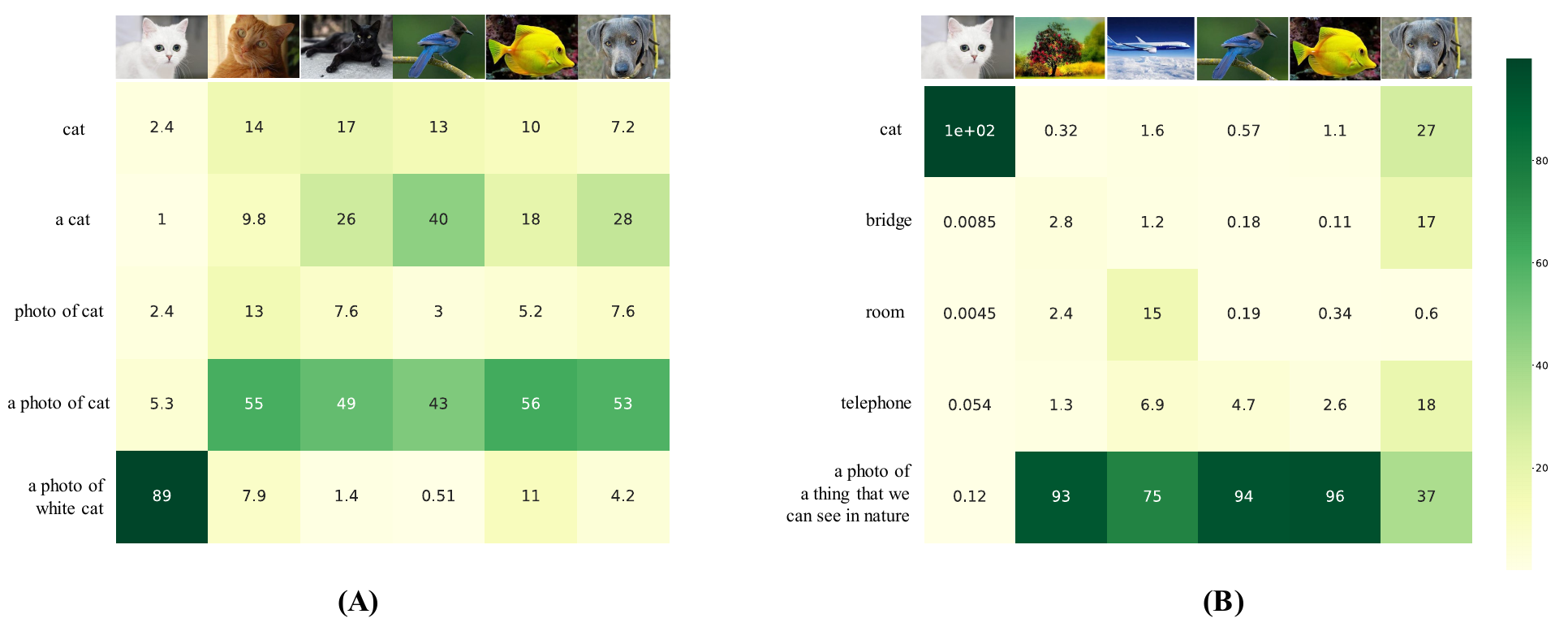}}

  \caption{Heatmaps depicting the cosine similarity between image inputs and ID concept vectors. In Figure (A), the ID concept vectors consist of sentences containing the word "cat" of varying lengths. It is observed that images tend to align with longer sentences regardless of whether there is a matching ID concept. Concurrently, keywords such as "white" significantly influence image matching. In Figure (B), aside from "cat", no ID concept can be precisely matched with the given images. However, other than cat images, all images exhibit a preference for aligning with a long sentence devoid of tangible objects. Notably, cat image remains unaffected, aligning solely with the ID concept "cat". All the data in the figure is obtained from the practical use of CLIP (https://github.com/openai/CLIP).}
  \label{fig:examples}
  
\end{figure*}

Although the employment of the Agent paradigm in the realm pertaining to LLMs has become a matter of course, the challenge lies in its adaptation to the domain of Out-of-distribution Detection. To address this issue, we have conducted a systematic examination of linguistic visual representations and, based on this, conducted targeted experiments, identifying three primary phenomena.Figure ~\ref{fig:examples} shows two basic examples. These laws provide a new perspective for us to understand and optimize multimodal learning models:

\begin{itemize}
      \item \textbf{The length of prompt will affect the prediction.} In similar text descriptions, the longer the prompt is within a certain range, the higher the score.\\
      Let $ L $ denote the length of the prompt, and $ S(L) $ denote the matching score. This relationship could be described as follows:

    $$
    S(L) \propto L \quad \text{for } L \in [a, b],
    $$
    where $ a $ and $ b $ denote the lower and upper bounds of the prompt length, respectively.
      \item \textbf{Different words have different weights in the text description.} The more important the word is in describing the overall image, the higher its weight. This means that keywords such as color and shape have a significant impact on the matching score.\\
      Let $ w_i $ denote the weight of the $ i $-th word, and let $ P $ denote the set of all words in the textual description. The overall weight of the description can be expressed as follows:

$$
W(P) = \sum_{i=1}^{n} w_i \cdot f_i(P)
$$
where $ f_i(P) $ is a function associated with the $ i $-th word, encompassing attributes such as word frequency, significance, and so forth, and $ n $ denotes the total number of words in the description.
      \item \textbf{Neutral prompts have little impact on images in the ID category.} Conversely, for images in the OOD category, neutral prompts can significantly reduce the score of the ID textual description.\\
      Let $ID$ denote the set of images in the ID category, $OOD$ denote the set of images in the OOD category, $N$ denote the set of neutral prompts, and $\Delta S(i, p)$ denote the change in score for an image $i$ when a prompt $p$ is applied. The following relationships hold:

$$
\forall i \in ID, \forall n \in N: |\Delta S(i, n)| \approx 0
$$
$$
\forall i \in OOD, \forall n \in N: \Delta S(i, n) \ll 0
$$
where $ |\Delta S(i, n)|\approx 0 $ indicates that the change in score for images in the ID category remains approximately zero when neutral prompts are applied, and $ \Delta S(i, n) \ll 0 $ indicates that the change in score for images in the OOD category is significantly negative when neutral prompts are applied.
\end{itemize}

The causes of these phenomena can be contemplated through the training methods and mechanisms of the CLIP-like models.Given that the common text input for pre-trained models during training is long sentences, they possess higher confidence in inferring long sentences. Additionally, CLIP-liked models undergo contrastive learning training, which enables the model to autonomously focus on the differences between various text descriptions, thereby giving higher weights to key words and skewing the predicted outcomes towards the positive class without being influenced by other words. We conducted a more detailed discussion and statistical verification in the \textbf{Appendix}.

Drawing on these significant observations and analyses, we propose the inclusion of neutral prompts unrelated to the ID category as Agents. By engaging Agents in a `collision' with OOD and ID data, we aim to distance the OOD data from the ID domain.

\subsection{Proposed approach}

Based on the aforementioned analysis, Concept Matching with Agent (CMA) employs neutral prompts as Agents to widen the gap between ID and OOD by constructing a vector triangle relationship. Essentially, CMA employs external agents to minimize the maximum score, which is fundamentally distinct from traditional learning methods. Therefore, it does not require additional data or training. Specifically, for a set of textual descriptions $t_i$, where $ i \in \{ 1, 2, ..., N\}$ corresponding to ID categories $\mathcal{Y}_{in}$, we select a certain number of neutral prompts $\mathcal{Y}_{ntc}$ to concatenate with them. The number is generally the same as the number of ID categories $N$, and the concatenated textual concepts $\mathcal{Y}_{con}$ are used as the input of the text encoder $\mathcal{T}$ to output text features $\mathbf{c}_i = \mathcal{T}(t_{i}) $, where $i \in \{1,2,...,2N \}$, $t_i \in \mathcal{Y}_{con} = \{ \mathcal{Y}_{in}, \mathcal{Y}_{ntc} \}$. Then, it is calculated with the image features output $\mathbf{v}' = \mathcal{I}(\mathbf{x}^{\prime})$ by the image encoder $\mathcal{I}$.

Formally, we define the \textbf{CMA} score as:

$$
\mathcal{S}_{\text{CMA}}(\mathbf{x}^{\prime}; \mathcal{Y}_{in}, \mathcal{Y}_{ntc}, \mathcal{T}, \mathcal{I}) = \frac{\exp(\text{sim}(\mathbf{v}', \mathbf{c}_{\hat{y}})/\tau)}{\sum_{i=1}^{2N}\exp(\text{sim}(\mathbf{v}', \mathbf{c}_i)/\tau)},
$$
where $\text{sim}(\cdot, \cdot)$ denotes cosine similarity, $\hat{y} = \mathop{\arg\max}\limits_{1 \leq i \leq N} \text{sim}(\mathbf{v}', \mathbf{c}_i)$ denotes the most matching concept index in ID categories, and $\tau$ is a temperature parameter used to adjust the distribution of similarity.

It is particularly noteworthy that although our calculation method is similar to MCM, there are significant differences in detail: (1) When calculating the highest score, the scores corresponding to neutral prompts should be excluded from the calculation, but when performing softmax scaling, the scores corresponding to neutral prompts should be retained; (2) For text descriptions of ID categories, we do not apply prompts for text enhancement, but only use the corresponding category names. This is to further differentiate between ID and neutral prompts, making the model more focused on whether the image truly corresponds to the ID category.

Finally, our OOD detection function can be formally formulated as:

$$
g\left(\mathbf{x}^{\prime} ; \mathcal{Y}_{\mathrm{in}}, \mathcal{Y}_{ntc}, \mathcal{T}, \mathcal{I}\right)=\left\{\begin{array}{ll}
1, & S_{\mathrm{CMA}} \geq \lambda \\
0, & otherwise
\end{array},\right.
$$
where 1 indicates ID and 0 indicates OOD. $\lambda$ is the threshold, and examples below $\lambda$ are considered OOD inputs.

\section{Experiments}
\label{sec:exp}

\begin{table*}[t]
\centering
\footnotesize
\caption{\textbf{Comparison results on ImageNet-1k OOD benchmarks.} We use ImageNet-1k as ID dataset. All methods use CLIP-B/16 as a backbone. Bold values represent the highest performance. }
\label{table:comparison_1k}
\footnotesize
\centering
     {\resizebox{\textwidth}{!}{
    \begin{tabular}{@{}lcccccccccccccc@{}} \toprule
    & \multicolumn{2}{c}{iNaturalist} && \multicolumn{2}{c}{SUN} && \multicolumn{2}{c}{Places} && \multicolumn{2}{c}{Texture} && \multicolumn{2}{c}{\textbf{Average}} \\ 
    \cmidrule(lr){2-3} \cmidrule(lr){5-6} \cmidrule(lr){8-9} \cmidrule(lr){11-12} \cmidrule(lr){14-15}
   \textbf{Method} & \scriptsize FPR95$\downarrow$ & \scriptsize AUROC$\uparrow$ && \scriptsize FPR95$\downarrow$ & \scriptsize AUROC$\uparrow$ && \scriptsize FPR95$\downarrow$ & \scriptsize AUROC$\uparrow$ && \scriptsize FPR95$\downarrow$ & \scriptsize AUROC$\uparrow$ && \scriptsize FPR95$\downarrow$ & \scriptsize AUROC$\uparrow$  \\ 
   
    \midrule
    &\multicolumn{12}{c}{\textbf{Requires training (or w. fine-tuning)}}          \\
    MSP \citep{hendrycks2016baseline} & 40.89 & 88.63 && 65.81 & 81.24 && 67.90 & 80.14 && 64.96 & 78.16 && 59.89 & 82.04 \\
    Energy \citep{liu2020energy} & 21.59 & 95.99 && 34.28 & 93.15 && 36.64 & 91.82 && 51.18 & 88.09 && 35.92 & 92.26 \\
    ODIN \citep{liang2017enhancing} & 30.22 & 94.65 && 54.04 & 87.17 && 55.06 & 85.54 && 51.67 & 87.85 && 47.75 & 88.80 \\
    ViM \citep{wang2022vim}& 32.19 & 93.16 && 54.01 & 87.19 && 60.67 & 83.75 && 53.94 & 87.18 && 50.20 & 87.82\\
    KNN \citep{sun2022out}& 29.17 & 94.52 && 35.62 & 92.67 && 39.61 & 91.02 && 64.35 & 85.67 && 42.19 & 90.97 \\
    NPOS \citep{tao2022non}& 16.58 & 96.19 && 43.77 & 90.44 && 45.27 & 89.44 && \textbf{46.12} & 88.80 && 37.93 & 91.22 \\
    CoOp \citep{zhou2022learning} & 43.38 & 91.26 && 38.53 & 91.95 && 46.68 & 89.09 && 50.64 & 87.83 && 44.81 & 90.03 \\
    LoCoOp \citep{LoCoOp} & 38.49 & 92.49 && 33.27 & 93.67 && 39.23 & 91.07 && 49.25 & \textbf{89.13} && 40.17 & 91.53\\
    \midrule
   &\multicolumn{12}{c}{\textbf{Zero-shot (no training required)}}   \\
    MCM \citep{MCM} & 30.94 & 94.61 && 37.67 & 92.56 && 44.76 & 89.76 && 57.91 & 86.10 && 42.82 & 90.76  \\ 
    GL-MCM  \citep{glmcm} & \textbf{15.18} & 96.71 && 30.42 & 93.09 && 38.85 & 89.90 && 57.93 & 83.63 && 35.47 & 90.83 \\
    CMA(Ours) & 23.84 & \textbf{96.89} && \textbf{30.11} & \textbf{93.69} &&	 \textbf{29.86} & \textbf{93.17} && 47.35 & 88.47 && \textbf{32.79} & \textbf{93.05} \\
    \bottomrule
    \end{tabular}
    }
    }


\centering
\footnotesize
\caption{\small Zero-shot OOD detection with CMA based on CLIP-B/16 with various ID datasets.}
\label{table:dif_id_datasets}
\footnotesize
\centering
     {\resizebox{\textwidth}{!}{
    \begin{tabular}{@{}lcccccccccccccc@{}} \toprule
    & \multicolumn{2}{c}{iNaturalist} && \multicolumn{2}{c}{SUN} && \multicolumn{2}{c}{Places} && \multicolumn{2}{c}{Texture} && \multicolumn{2}{c}{\textbf{Average}} \\ 
    \cmidrule(lr){2-3} \cmidrule(lr){5-6} \cmidrule(lr){8-9} \cmidrule(lr){11-12} \cmidrule(lr){14-15}
   \textbf{ID datasets} & \scriptsize FPR95$\downarrow$ & \scriptsize AUROC$\uparrow$ && \scriptsize FPR95$\downarrow$ & \scriptsize AUROC$\uparrow$ && \scriptsize FPR95$\downarrow$ & \scriptsize AUROC$\uparrow$ && \scriptsize FPR95$\downarrow$ & \scriptsize AUROC$\uparrow$ && \scriptsize FPR95$\downarrow$ & \scriptsize AUROC$\uparrow$  \\ 
   
    \midrule
    FashionMNIST \citep{fashion-mnist} & 0.00 & 100.00 && 0.00 & 100.00 && 0.00 & 100.00 && 13.61 & 93.60 && 3.40 & 98.40 \\
    STL10 \citep{stl10} & 0.00 & 100.00 && 2.56 & 99.01 && 0.00 & 100.00 && 0.00 & 100.00 && 0.64 & 99.75 \\
    OxfordIIIPet \citep{pet} & 0.00 & 99.89 && 0.37 & 99.89 && 0.96 & 99.71 && 0.35 & 99.85 && 0.42 & 99.84 \\
    Food101 \citep{food101} & 0.25 & 99.90 && 0.49 & 99.86 && 1.79 & 99.40 && 2.99 & 99.33 && 1.38 & 99.62 \\
    CUB-200 \citep{cub200} & 0.00 & 100.00 && 0.00 & 100.00 && 0.00 & 99.99 && 0.00 & 100.00 && 0.00 & 100.00 \\
    PlantVillage \citep{PlantVillage} & 2.95 & 97.83 && 0.18 & 98.11 && 1.12 & 98.41 && 4.37 & 97.56 && 2.16 & 97.98 \\
    LFW \citep{lfw} & 2.89 & 99.14 && 2.24 & 99.52 && 8.04 & 98.25 && 18.88 & 95.34 && 8.01 & 98.06 \\
    Stanford-dogs \citep{dogs} & 0.11 & 99.92 && 0.16 & 99.91 && 0.58 & 99.73 && 0.67 & 99.72 && 0.38 & 99.82 \\
    FGVC-Aircraft \citep{fgvc} & 0.00 & 99.99 && 0.67 & 99.87 && 1.02 & 99.69 && 0.00 & 99.99 && 0.42 & 99.89 \\
    Grocery Store \citep{grocery} & 0.03 & 99.89 && 0.15 & 99.97 && 0.69 & 99.82 && 0.79 & 99.76 && 0.42 & 99.86 \\
    CIFAR10 \citep{cifar}& 0.00 & 100.00 && 5.12 & 98.29 && 2.56 & 99.67 && 0.00 & 99.88 && 1.92 & 99.46 \\
    CIFAR100 \citep{cifar} & 12.80 & 97.15 && 17.92 & 95.66 && 15.36 & 96.19 && 4.53 & 98.42 && 12.65 & 96.86 \\
    \bottomrule
    \end{tabular}
    }
    }

\end{table*}

\subsection{Setup}

\paragraph{Datasets} 
We conducted a comprehensive evaluation of the performance of our method across various dimensions and compared it with widely employed OOD detection algorithms. (1) We assessed our approach on the ImageNet-1k OOD benchmark. This benchmark utilizes the large-scale visual dataset ImageNet-1k \citep{imagenet1k} as the ID data and four OOD datasets (including subsets of iNaturalist \citep{van2018inaturalist}, SUN \citep{xiao2010sun}, Places\citep{zhou2017places}, and Textures \citep{textures}, which are same as Sun \textit{et al.} \citep{sun2022out}) to fully evaluate the method's performance across various semantic and scenario contexts. (2) We evaluated our method on various small-scale datasets. Specifically, we considered the following ID datasets: FashionMNIST \citep{fashion-mnist}, STL10 \citep{stl10}, OxfordIIIPet \citep{pet}, Food-101 \citep{food101}, CUB-200 \citep{cub200}, PlantVillage \citep{PlantVillage}, LFW \citep{lfw}, Stanford-dogs \citep{dogs}, FGVC-Aircraft \citep{fgvc}, Grocery Store \citep{grocery}, and CIFAR-10 \citep{cifar}. (3) We assessed our method on hard OOD tasks \citep{semantically_hard_ood, spurious_OOD}. Following the standards of the MCM \citep{MCM}, we evaluated using subsets of ImageNet-1k, namely ImageNet-10 and ImageNet-20, which have similar classes (e.g., dog (ID) vs. wolf (OOD)). During the experiments, we ensured that each OOD dataset did not overlap with the ID dataset in terms of classes.

\paragraph{Model}
In our experiment, all algorithms uniformly employ CLIP \citep{CLIP} as the pre-trained model, which is one of the most prevalent and publicly available visual-linguistic models. Specifically, we utilize CLIP-B/16 as the foundational evaluation model, consisting of a ViT-B/16 transformer \citep{vit} serving as the image encoder and a masked self-attention Transformer \citep{transformer} as the text encoder. Additionally, unless otherwise specified, the temperature coefficient is uniformly set to 1 across all algorithms.

\paragraph{Metrics}
For evaluation, we use the following metrics: (1) the false positive rate (FPR95) of OOD samples when the true positive rate of in-distribution samples is at 95\%, (2) the area under the receiver operating characteristic curve (AUROC). All evaluation outcomes for our method are derived from the average of three experiments, and (3) ID classification accuracy (ID ACC).

\subsection{Main Results}

\paragraph{OOD detection on Large-scale datasets.}
The benchmarking of large-scale OOD datasets demonstrates the feasibility of the method for real-world applications and holds significant value. Typically, we employ ImageNet-1k as the ID dataset for Large-scale OOD detection. Table ~\ref{table:comparison_1k} presents the performance of our method in comparison with other approaches under this benchmark. Overall, our method surpasses other methods, achieving superior performance. When juxtaposed against the average performance of Zero-shot methods, CMA demonstrates enhancements of \textbf{2.22\%} in terms of AUROC and \textbf{2.68\%} in terms of FPR95. Similarly, when juxtaposed against the average performance of Require training methods, CMA also demonstrates improvements, achieving enhancements of \textbf{0.79\%} in AUROC and \textbf{3.13\%} in FPR95, thereby showcasing its exceptional performance.

\paragraph{OOD detection on small-scale datasets.}
In contrast to benchmarks based on large-scale OOD datasets, small-scale OOD detection often features fewer distinct ID categories. Demonstrating robust performance across these categories is indicative of a method's scalability. Table ~\ref{table:dif_id_datasets} illustrates the efficacy of our approach across various ID datasets. A notable outcome is that our method achieves impressive performance across these datasets, especially when no specific training was tailored to each dataset.

\section{Discussion}
\label{sec:discussion}
\begin{figure*}[t]
  \centering
  \includegraphics[width=0.75\linewidth]{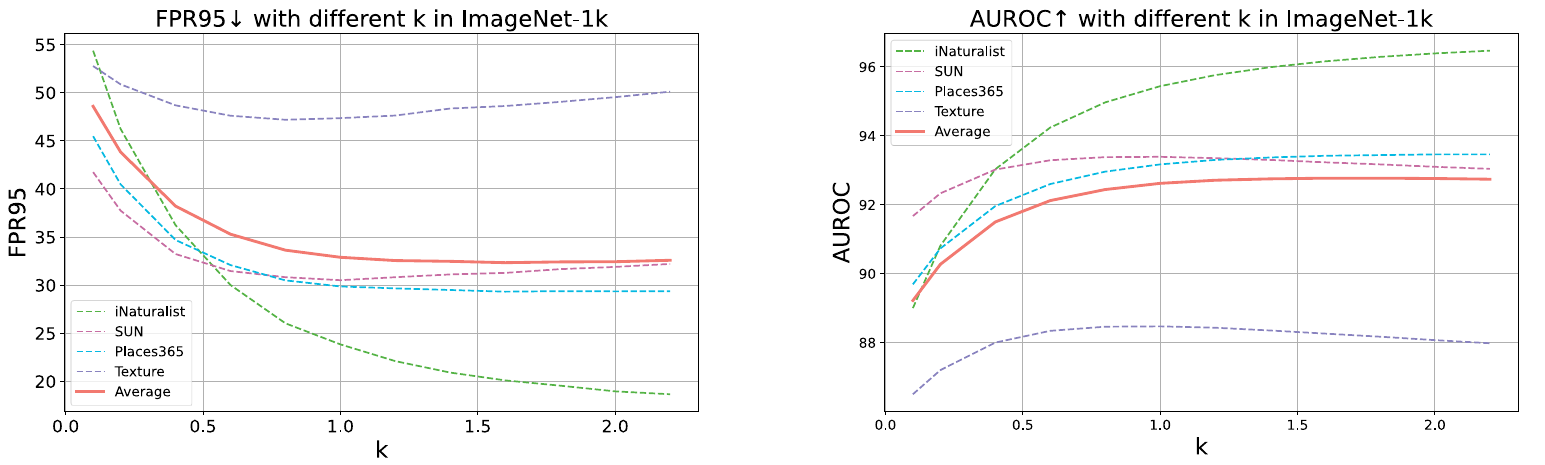}
  \caption{\small \textbf{Impact curve of} $k = \frac{\text{number of Agents}}{\text{number of ID Labels}}$ \textbf{on the performance of CMA.} \textit{Left} shows that as $k$ gradually increases, the FPR95 on various datasets generally decreases, with the fastest decline occurring in the range of $k$ less than 0.5, followed by a gradual slowdown, which is more evident on the average curve. \textit{Right} shows that as $k$ gradually increases, the AUROC gradually increases, also with a rapid rise followed by a gradual slowdown.}
  \label{fig:FPR95_AUROC_k}
\end{figure*}
\begin{figure*}[t]
  \centering
  \includegraphics[width=0.75\linewidth]{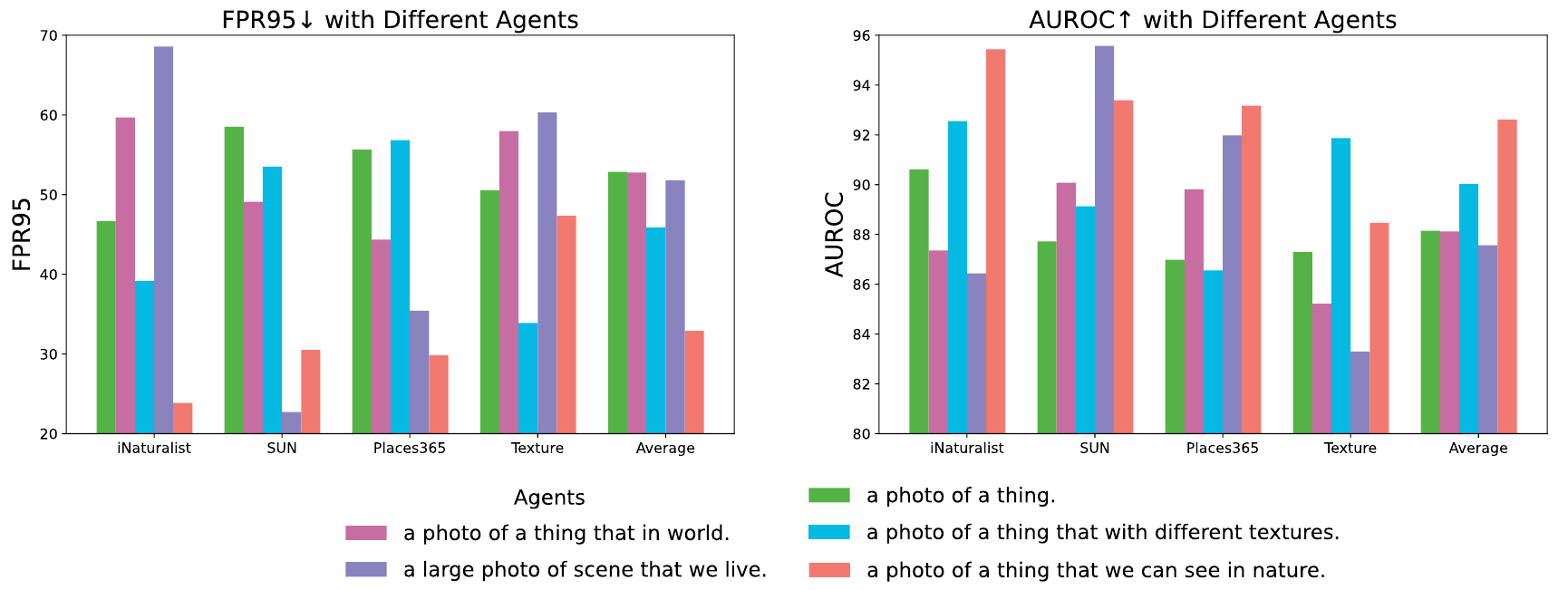}
  \caption{\small \textbf{Comparison with different Agents.} \textit{Left} shows the performance of different agents on various datasets in terms of FPR95, while \textit{Right} shows the performance in terms of AUROC. Clearly, the agent that performs best on average across all datasets does not perform best on every dataset. Moreover, even agents that perform poorly on average can still show decent performance on certain datasets.}
  \label{fig:FPR95_AUROC_agents}
\end{figure*}

\paragraph{The number of Agents.}

In section~\ref{sec:method}, we present the score calculation formula $\mathcal{S}_{\text{CMA}}(\mathbf{x}^{\prime}; \mathcal{Y}_{in}, \mathcal{Y}_{ntc}, \mathcal{T}, \mathcal{I})$ for CMA, where we set the number of Agents to $N$, aligning with the number of ID categories. This number was determined based on extensive experiments, with $k$ set to $k=\frac{M}{N}$, where $M$ is the number of Agents.Concurrently, the formula for calculating the score has been modified to 
$\mathcal{S}_{\text{CMA}}'(\mathbf{x}^{\prime}; \mathcal{Y}_{in}, \mathcal{Y}_{ntc}, \mathcal{T}, \mathcal{I}) = \frac{\exp(\text{sim}(\mathbf{v}', \mathbf{c}_{\hat{y}})/\tau)}{\sum_{i=1}^{N}\exp(\text{sim}(\mathbf{v}', \mathbf{c}_i)/\tau) + \sum_{j=1}^{M}\exp(\text{sim}(\mathbf{v}', \mathbf{c}_{j+N})/\tau)}$, where $\hat{y} = \mathop{\arg\max}\limits_{1 \leq i \leq N} \text{sim}(\mathbf{v}', \mathbf{c}_i)$. Experiments were conducted for various values of $k$. Figure \ref{fig:FPR95_AUROC_k} illustrates the results of our experiments conducted on ImageNet-1k. The optimal value of $k$ was identified as 1. Additionally, it was observed that an excessive number of Agents does not enhance the overall performance of CMA, whereas a insufficient number of Agents can adversely affect it. This finding illuminates the rationality of the vector triangle relationship. An insufficient number of Agents fails to exert the same impact on OOD data as it does on ID categories, leading to structural disarray and performance degradation. Once the number of Agents reaches a certain threshold, a stable structure is established, after which increasing the number of Agents does not significantly improve performance.

\paragraph{Performance of different Agents.}

Figure \ref{fig:FPR95_AUROC_agents} illustrates the impact of different neutral terms on the OOD detection performance of agents. It is evident that the impact of different agents on performance is substantial. On average, the best and worst outcomes in terms of FPR95 and AUROC can differ by \textbf{19.95\%} and \textbf{4.50\%}, respectively. From another perspective, certain agents excel in specific contexts, such as "a large photo of the scene we live in." While their overall performance across multiple environments lags significantly behind the optimal level, they achieve their best results when using the SUN dataset \citep{xiao2010sun} as OOD data. On one hand, this highlights a potential limitation of CMA, where different agents can yield disparate outcomes. On the other hand, it underscores the significant potential of CMA, as it can tailor specific agents to suit diverse environments, thereby enhancing performance.

\section{Conclusion}
\label{sec:conclusion}

This paper proposes a novel zero-shot OOD detection framework, Concept Matching with Agent (CMA). By introducing the concept of Agents into the OOD detection task, a vector triangular relationship consisting of ID labels, data inputs, and Agents is constructed, offering a fresh perspective on OOD detection. Beginning with a language-vision representation, we demonstrate the impact of neutral words on CLIP-liked models, thereby proposing the innovative idea of treating neutral words as Agents. Building on this, we propose a novel score computation method based on CMA. By incorporating the relationships between Agents and data inputs, this method enables unique interactions between ID data and OOD data with Agents, thereby facilitating a better separation between ID and OOD. We investigate the effectiveness of CMA across various scenarios, including large-scale datasets, small-scale datasets, and hard OOD detection, achieving outstanding performance across a wide range of tasks. Finally, we delve deeper into CMA, highlighting its flexibility and scalability. We hope that our work will inspire future exploration of new paradigms for OOD detection.

\section*{Acknowledgements}
This work was supported by National Natural Science Foundation of China, Grant Number:
62476109, 62206108, 62176184, and the Natural Science Foundation of Jilin Province, Grant Number: 20240101373JC, and Jilin Province Budgetary Capital Construction Fund Plan, Grant Number: 2024C008-5, and Research Project of Jilin Provincial Education Department, Grant Number: JJKH20241285KJ.

\bibliography{aaai25}

\newpage
\clearpage
\appendix

\begin{center}
\textbf{\Large Appendix}
\end{center}







\section{Cause Analysis and Statistical Analysis of Three Phenomena}

\subsection{Impact of prompt length on matching score}
\begin{figure}[htb]
  \centering
  \includegraphics[width=0.45 \textwidth]{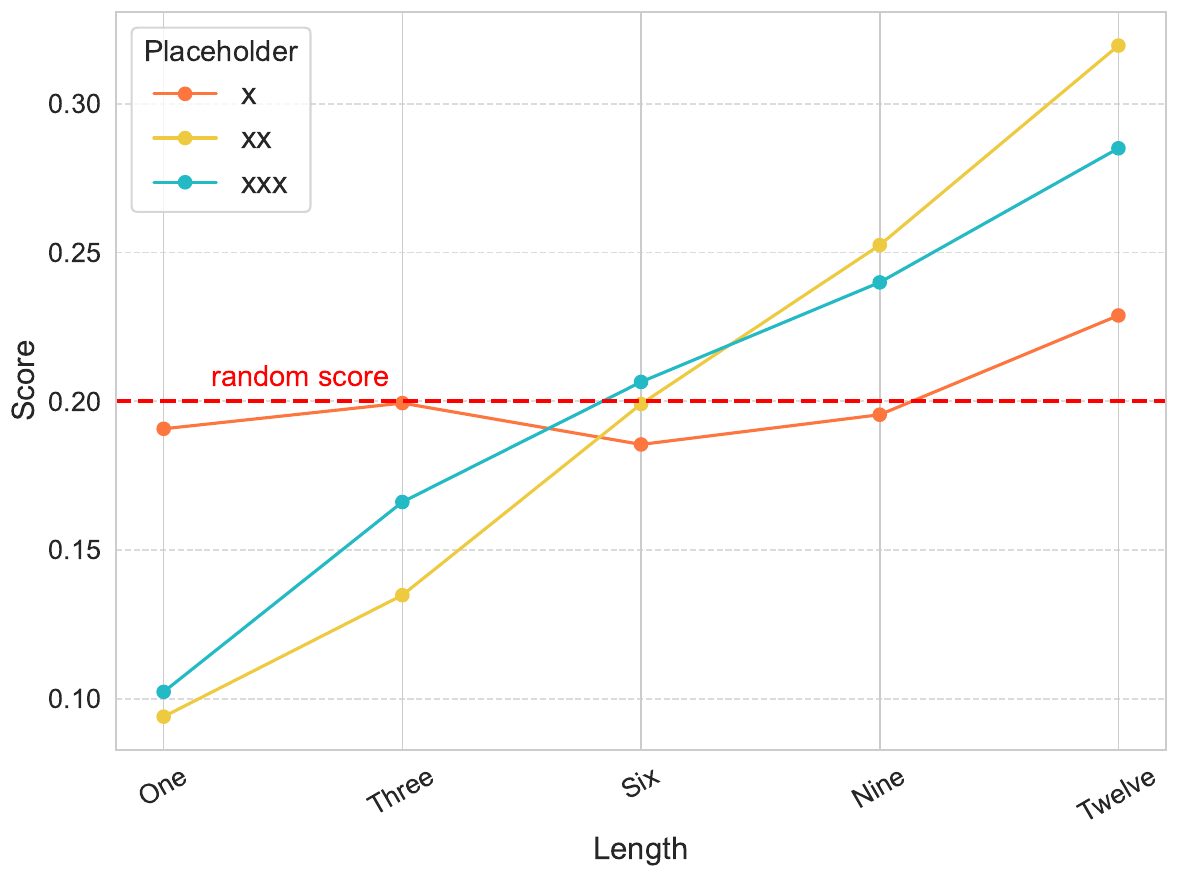}
  \caption{Distribution of scores for prompts of different lengths with different placeholders}
  \label{fig:evi1_1}
\end{figure}

It is challenging to directly confirm the relationship between prompt length and matching score, as the matching score is obviously affected by multiple factors. A feasible method is to use only placeholders as prompts, such as "xx xx", which reduces the influence of the prompt itself and focuses on the length of the prompt. Therefore, we conducted a statistical experiment (Figure \ref{fig:evi1_1}) using the CIFAR-10 as image inputs, with different lengths of the same placeholder as prompt inputs, and then averaged all the scores to obtain a statistical score distribution. The results clearly indicate a positive correlation between prompt length and matching score.

Based on the above understanding, we define $S$ as a function of $L$, denoted as $S(L)$, and assume that within the range $R$, $S(L)$ is a monotonically increasing function. A simple linear model is proposed:

\begin{equation*}
S(L) = aL + b, \quad \text{for } L \in R
\end{equation*}

where $a > 0$ ensures that the function is increasing. To validate this relationship statistically, we employ a linear regression model using a dataset of $n$ prompts and their corresponding scores:

\begin{equation*}
S_i = \beta_0 + \beta_1L_i + \varepsilon_i, \quad i = 1, 2, \ldots, n
\end{equation*}

where $\beta_0$ is the intercept, $\beta_1$ is the slope coefficient representing the effect of prompt length on the prediction score, and $\varepsilon_i$ is the error term. The significance of $\beta_1$ is tested using a t-test:

\begin{equation*}
t = \frac{\hat{\beta_1}}{SE(\hat{\beta_1})}
\end{equation*}

where $\hat{\beta_1}$ is the estimated value of $\beta_1$ and $SE(\hat{\beta_1})$ is its standard error. If $|t| > t_\text{crit}$, where $t_\text{crit}$ is the critical value from the t-distribution at a chosen significance level, we reject the null hypothesis ($H_0: \beta_1 = 0$) and conclude that prompt length significantly affects the prediction score. Furthermore, a positive $\beta_1$ would support our hypothesis of a positive relationship between prompt length and prediction performance. 

\begin{figure}[htb]
  \centering
  \includegraphics[width=0.45 \textwidth]{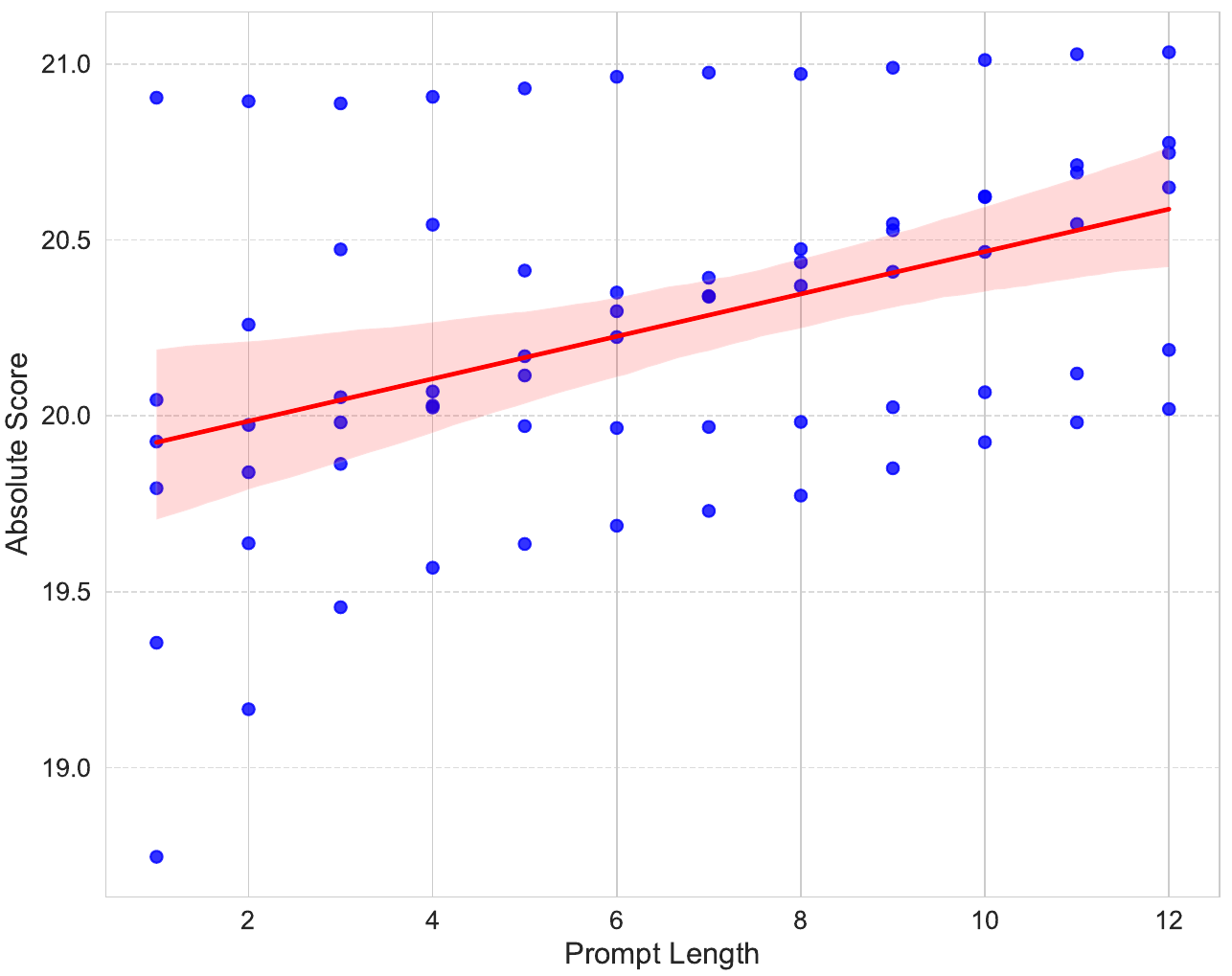}
  \caption{T-test results on the distribution of Absolute scores for prompts of different lengths}
  \label{fig:evi1_2}
\end{figure}

We conducted more extensive experiments on the larger dataset ImageNet-1k, selecting six placeholders "x", "xx", "xxx", "thing", "time", and "the". We removed softmax scaling and only performed more direct similarity matching. The matching results were called "Absolute Score", and then t-test was performed. The results (Figure \ref{fig:evi1_2}) showed that "\textbf{Length significantly affects Score}".

\subsection{Weight of words in text descriptions}
\begin{figure}[htb]
  \centering
  \includegraphics[width=0.45 \textwidth]{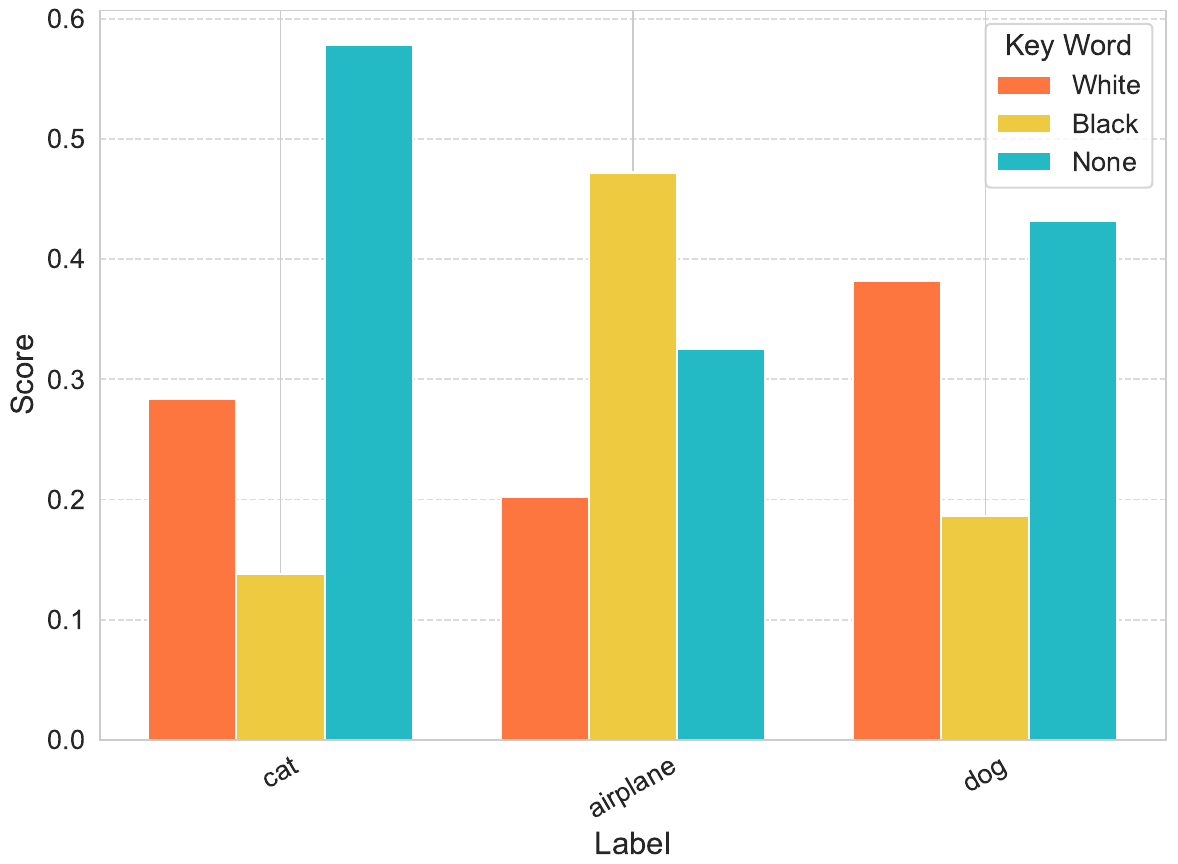}
  \caption{Scores on different keywords}
  \label{fig:evi2_1}
\end{figure}

We conducted statistical experiments on specific categories of CIFAR-10 using different prompts (Figure \ref{fig:evi2_1}). For example, we used "white cat", "black cat", and "a cat" as prompts, and images labeled as "cat" in CIFAR-10 as image inputs to obtain a large number of matching scores, and finally averaged them to obtain a score distribution. It can be seen that although all three prompts represent "cat", and "a something" should have stronger universality, there are still a large number of inputs that tend to be "white" and "black", indicating the important influence of key words on matching scores.

Let $P$ represent the set of all words in a given textual description, and $W_i$ denote the weight of the $i$-th word. We propose that the overall weight of the description, $W(P)$, can be expressed as a function of individual word weights and their associated attributes:

$$
W(P) = \sum_{i=1}^{n} W_i \cdot f_i(P)
$$

where $n$ is the total number of words in the description, and $f_i(P)$ is a multifaceted function associated with the $i$-th word. This function encapsulates various linguistic and contextual attributes, including but not limited to:

\begin{itemize}
    \item Word frequency: The occurrence rate of the word within the description and in a broader corpus.
    \item Term frequency-inverse document frequency (TF-IDF): A numerical statistic reflecting the word's importance in the document relative to a collection of documents.
    \item Semantic significance: The word's relevance to the overall meaning or topic of the description.
    \item Syntactic role: The word's grammatical function within the sentence structure.
    \item Position: The location of the word within the description, potentially indicating its importance.
\end{itemize}

\subsection{Impact of neutral prompts on images in ID and OOD inputs}
\begin{figure}[htb]
  \centering
  \includegraphics[width=0.45 \textwidth]{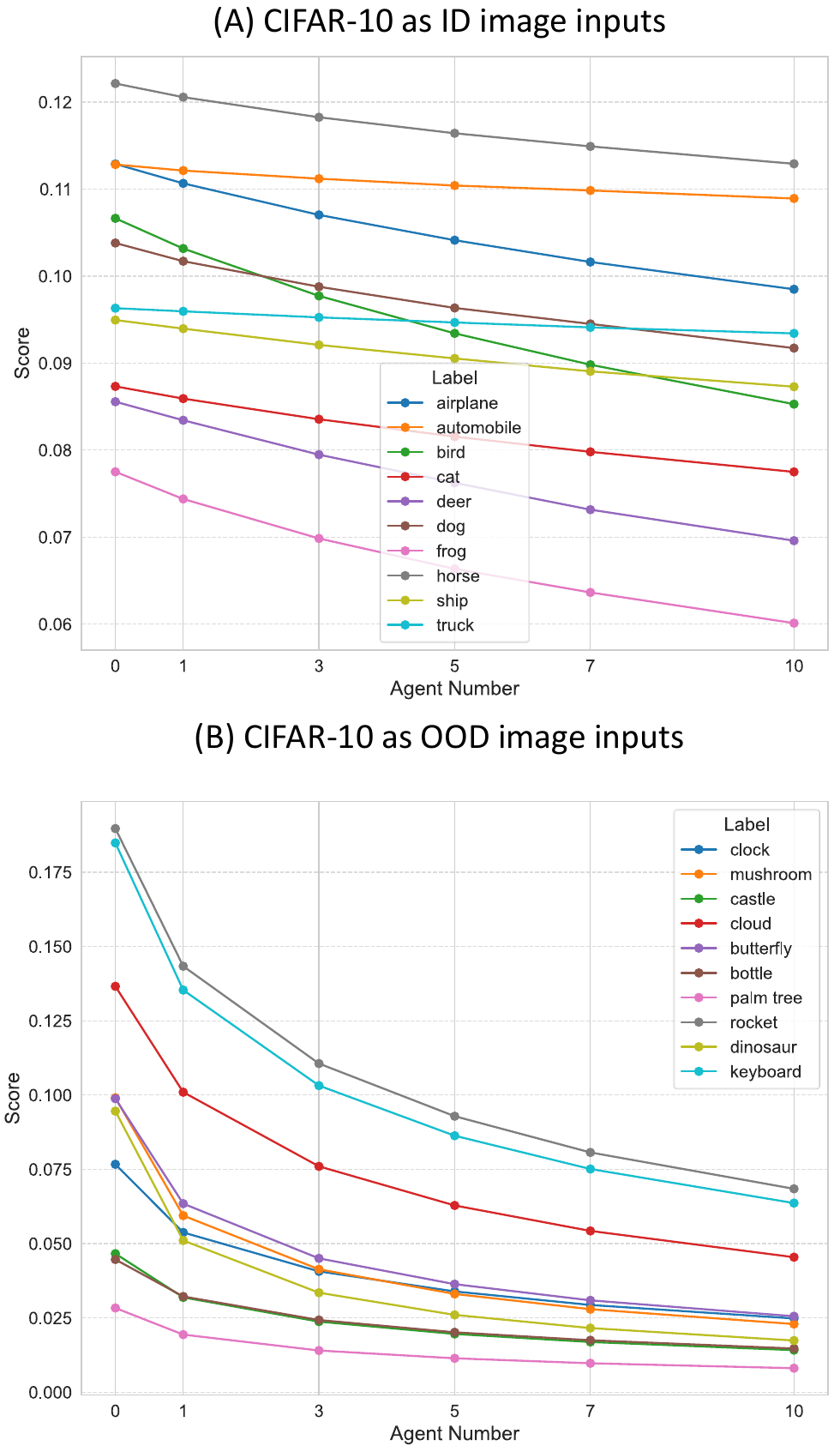}
  \caption{Score of different labels with different agent number}
  \label{fig:evi3}
\end{figure}

Verifying the minimal impact of neutral prompt words on ID inputs is not difficult: We designed two sets of statistical experiments (Figure \ref{fig:evi3}), one using CIFAR-10 as ID inputs, where the prompts were the labels of CIFAR-10, and the other using CIFAR-10 as OOD inputs, where the prompts were labels similar to those of CIFAR-10 in CIFAR-100. By gradually increasing the number of agents (neutral prompts), we observed the trend of score distribution changes. The results are clear: when faced with OOD inputs, the scores for each label decrease rapidly as the number of agents increases, while for ID inputs, the decrease is slower.

Formally, let us define the following sets:

- $ID$: the set of in-distribution images, i.e., images that belong to the categories the labels include

- $OOD$: the set of out-of-distribution images, i.e., images from categories not included in the labels

- $N$: the set of neutral prompts, defined as prompts that do not contain specific information about image content

We introduce $\Delta S(i, p)$ as the change in recognition score for image $i$ when prompt $p$ is applied. Our hypothesis posits two key relationships:

For in-distribution images:
   $$\forall i \in ID, \forall n \in N: |\Delta S(i, n)| \leq \epsilon$$
   where $\epsilon$ is a small positive constant representing the threshold for negligible change.

For out-of-distribution images:
   $$\forall i \in OOD, \forall n \in N: \Delta S(i, n) < -\delta$$
   where $\delta$ is a positive constant representing a significant decrease in score.

These relationships can be formalized as:

$$
P(|\Delta S(i, n)| \leq \epsilon | i \in ID, n \in N) \geq 1 - \alpha
$$
$$
P(\Delta S(i, n) < -\delta | i \in OOD, n \in N) \geq 1 - \beta
$$
where $\alpha$ and $\beta$ are small positive constants representing the acceptable error rates for the respective hypotheses. This formulation allows for statistical testing of the hypotheses using empirical data.

\begin{figure}[htb]
  \centering
  \includegraphics[width=0.45 \textwidth]{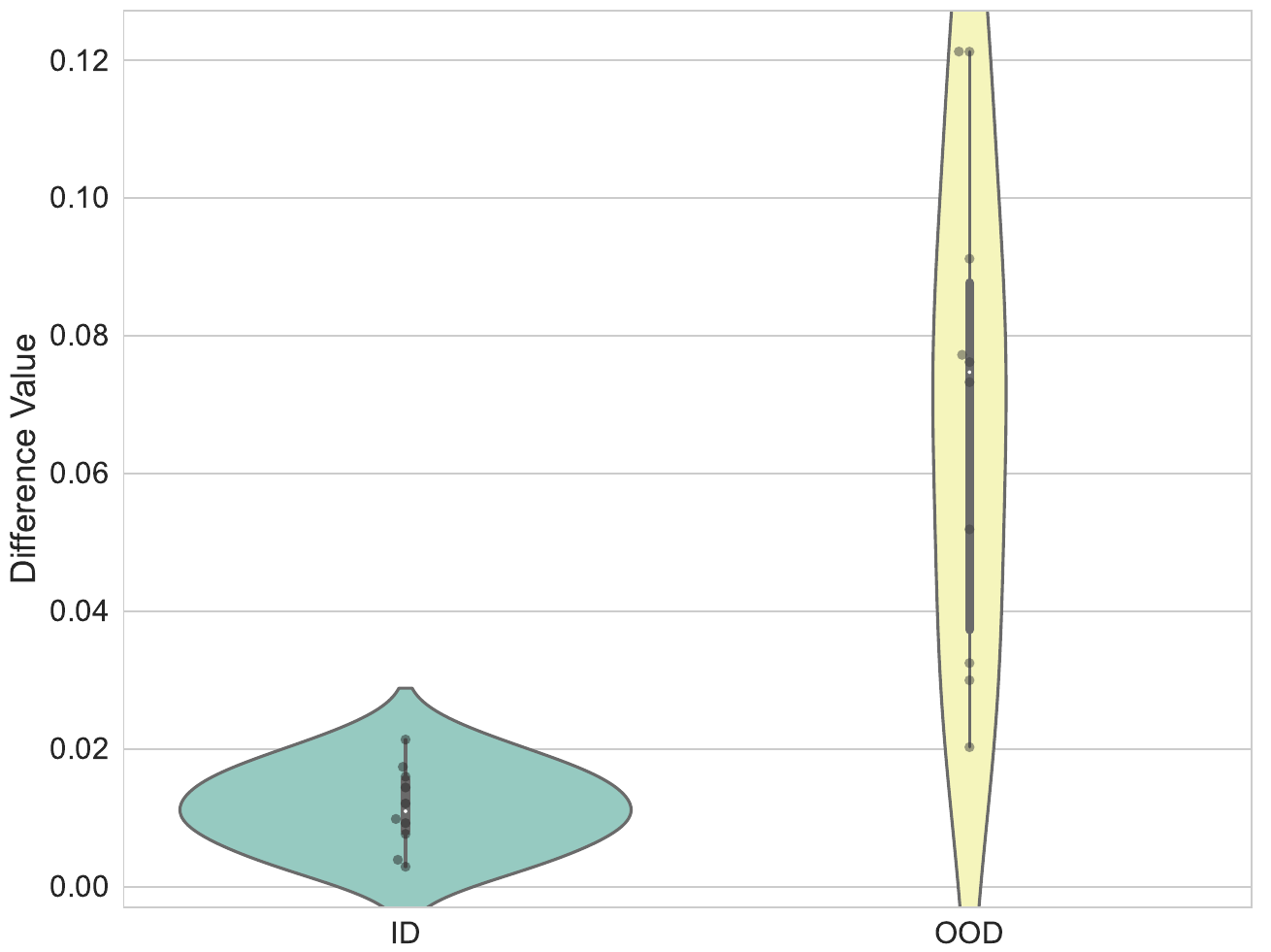}
  \caption{Comparison of difference distribution: ID inputs vs OOD inputs}
  \label{fig:evi3_5}
\end{figure}

This can be verified by calculating the distribution of the differences between ID inputs and OOD inputs with and without an agent (Figure \ref{fig:evi3_5}). Clearly, the distribution of differences for ID inputs is smaller and more concentrated, while that for OOD inputs is larger and more dispersed.

\section{Method for Obtaining Neutral Prompts}
The creation of neutral prompts through manual construction is feasible to a certain extent, by selecting words devoid of actual meaning to compose a short sentence. However, it is evident that employing large language models for the generation of neutral prompts is a superior option, as these models can generate a greater diversity of prompts, distinguishing themselves from manual constructions. We have developed two versions of prompts (See \ref{Prompt-1} and \ref{Prompt-2}) to generate neutral prompts by large language models, the former being more stable, producing similar neutral prompts, while the latter is more diverse, capable of generating a wider array of prompts. The neutral prompts used and mentioned in this paper and in the experiments are all obtained through manual selection after generation by the large language models.

\section{Near OOD Detection Task}

\begin{table*}[t]
\centering
\footnotesize
\caption{\small Comparison results on  \textbf{near OOD detection} tasks. }
\label{tab:hard}
\footnotesize
\resizebox{\textwidth}{!}{
\small
\begin{tabular}{llccccc}
\toprule
 \multirow{2}{*}{\textbf{Method}} & \textbf{ID} & ImageNet-10 & ImageNet-20 & ImageNet-10 & ImageNet-20 & Waterbirds \\
 & \textbf{OOD}        &  ImageNet-20 &  ImageNet-10 & ImageNet-100 & ImageNet-100 & Spurious OOD  \\
 \midrule
 && FPR95 / AUROC & FPR95 / AUROC & FPR95 / AUROC & FPR95 / AUROC & FPR95 / AUROC\\
MSP \citep{hendrycks2016baseline}  & & 9.38 / 98.31 & 12.51 / 97.70 & 5.82 / 98.88 & 14.14 / 95.73 & 39.57 / 90.99 \\
MCM \citep{MCM}   & & {5.00} / {98.71} & 12.91 / {98.09} & 3.70 / 99.09 & 13.16 / 96.32 & 5.87 / 98.36    \\
CMA(Ours) &    &  3.1 / 99.19&  6.2 / 98.71 & 2.94 / 99.18 & 10.32 / 96.64& 3.22 / 99.01\\
\bottomrule
\end{tabular}
}
\end{table*}

Table ~\ref{tab:hard} presents the outcomes of various zero-shot approaches in near out-of-distribution (OOD) detection. For dataset configurations, we alternated between Imagenet-10 and Imagenet-20 (See Appendix \ref{sec:ap_exp_detail}), which share similar semantic content, as the ID and OOD datasets, respectively. Alternatively, we employed the setup of using Imagenet-10 and Imagenet-20 as the ID datasets, with Imagenet-100 serving as the OOD dataset. Additionally, to account for spurious correlations, we utilized the Spurious OOD detection benchmark introduced by Ming \textit{et al.} \citep{MCM} and conducted comparative experiments on the Waterbirds dataset \citep{spurious_OOD}. Across various experiments, CMA outperformed MSP and MCM in terms of AUROC and FPR95.

\section{Ablation Study}

\subsection{Whether to use softmax scaling}

\begin{table*}[t]
\centering
\footnotesize
\caption{Comparison results of CMA with and without softmax scaling based on CLIP-B/16. The ID dataset is ImageNet-1k.}
\label{table:comparison_softmax}
\footnotesize
\centering
     {\resizebox{\textwidth}{!}{
    \begin{tabular}{@{}lcccccccccccccc@{}} \toprule
    & \multicolumn{2}{c}{iNaturalist} && \multicolumn{2}{c}{SUN} && \multicolumn{2}{c}{Places} && \multicolumn{2}{c}{Texture} && \multicolumn{2}{c}{\textbf{Average}} \\ 
    \cmidrule(lr){2-3} \cmidrule(lr){5-6} \cmidrule(lr){8-9} \cmidrule(lr){11-12} \cmidrule(lr){14-15}
   \textbf{Method} & \scriptsize FPR95$\downarrow$ & \scriptsize AUROC$\uparrow$ && \scriptsize FPR95$\downarrow$ & \scriptsize AUROC$\uparrow$ && \scriptsize FPR95$\downarrow$ & \scriptsize AUROC$\uparrow$ && \scriptsize FPR95$\downarrow$ & \scriptsize AUROC$\uparrow$ && \scriptsize FPR95$\downarrow$ & \scriptsize AUROC$\uparrow$  \\ 
   
    \midrule
    w/o softmax scaling & 70.27 & 85.64 && 57.75 & 88.49 && 46.31 & 89.80 && 80.67 & 77.75 && 63.75 & 85.42 \\
    w/ softmax scaling & 23.84 & 96.89 && 30.11 & 93.69 &&	 29.86 & 93.17 && 47.35 & 88.47 && 32.79 & 93.05 \\
    \bottomrule
    \end{tabular}
    }
    }


\centering
\footnotesize
\caption{Comparison results of CMA with different models. The ID dataset is ImageNet-1k.}
\label{table:comparison_models}
\footnotesize
\centering
     {\resizebox{\textwidth}{!}{
    \begin{tabular}{@{}lcccccccccccccc@{}} \toprule
    & \multicolumn{2}{c}{iNaturalist} && \multicolumn{2}{c}{SUN} && \multicolumn{2}{c}{Places} && \multicolumn{2}{c}{Texture} && \multicolumn{2}{c}{\textbf{Average}} \\ 
    \cmidrule(lr){2-3} \cmidrule(lr){5-6} \cmidrule(lr){8-9} \cmidrule(lr){11-12} \cmidrule(lr){14-15}
   \textbf{Method} & \scriptsize FPR95$\downarrow$ & \scriptsize AUROC$\uparrow$ && \scriptsize FPR95$\downarrow$ & \scriptsize AUROC$\uparrow$ && \scriptsize FPR95$\downarrow$ & \scriptsize AUROC$\uparrow$ && \scriptsize FPR95$\downarrow$ & \scriptsize AUROC$\uparrow$ && \scriptsize FPR95$\downarrow$ & \scriptsize AUROC$\uparrow$  \\ 
   
    \midrule
    CLIP-RN50 & 24.08 & 95.25 && 37.54 & 91.32 && 44.53 & 88.92 && 39.21 & 90.84 && 36.34 & 91.58 \\
    CLIP-B/16 & 23.84 & 96.89 && 30.11 & 93.69 &&	 29.86 & 93.17 && 47.35 & 88.47 && 32.79 & 93.05 \\
    CLIP-L/14 & 19.52 & 96.34 && 25.78 & 94.69 && 26.54 & 94.29 && 49.48 & 88.21 && 30.33 & 93.38 \\
    \bottomrule
    \end{tabular}
    }
    }

\end{table*}

Although MCM has demonstrated the significant role of softmax scaling in CLIP-based OOD detection, our method CMA underscores the even more crucial role of softmax scaling. Table ~\ref{table:comparison_softmax} illustrates the disparities between the use and non-use of softmax scaling. A crucial factor is that without softmax scaling, the score formula in CMA would become: $$\mathcal{S}_{\text{CMA}}^{w/o} = \text{sim}(\mathbf{v}', \mathbf{c}_{\hat{y}})/\tau,$$ where $\hat{y} = \mathop{\arg\max}\limits_{1 \leq i \leq N} \text{sim}(\mathbf{v}', \mathbf{c}_i)$ represents the most matching concept index in ID categories. In comparison to the original formula, the absence of softmax scaling essentially excludes Agents from the final score calculation, resulting in a score of $\mathcal{S}_{\text{CMA}}^{w/o} = \max_{i \in [N]} \text{sim}(\mathbf{v}', \mathbf{c}_{i})/\tau$. This prevents the construction of vector triangle relationships, thus negating the impact of Agents and leading to a significant reduction in performance. This also validates the rationality and effectiveness of $\mathcal{S}_{\text{CMA}}$.

\subsection{Comparative performance across models}
We conducted a comparative analysis of the performance of CMA across various models \citep{CLIP}, with the results presented in Table ~\ref{table:comparison_models}. The findings indicate that as the backbone model's performance improves, the effectiveness of CMA also intensifies, resulting in superior OOD detection capabilities. Furthermore, this study underscores the fact that our approach is not reliant on a single model, but can be effectively integrated across diverse models, demonstrating remarkable scalability.

\section{Related Works}
\label{sec:relate_works}

\paragraph{Single-modal Out-of-distribution detection}
Visual Out-of-distribution (OOD) detection \citep{survey1, survey2, yang2022openood}, a significant area in machine learning systems and computer vision, involves discerning images devoid of specific semantic information. Following a prolonged period of development, this field has evolved into two broad categories: single-modality OOD detection methods and multimodal OOD detection approaches that leverage semantic information. For the former, D Hendrycks \textit{et al.} \citep{hendrycks2016baseline} outlines the fundamental paradigm within neural networks, which has been the foundation for numerous subsequent developments, including output-based methods \citep{bendale2016towards,devries2018learning,du2022vos,hein2019relu,hsu2020generalized,huang2021mos,liu2020energy,sun2022dice}, density-based methods \citep{abati2019latent,zong2018deep,deecke2019image,sabokrou2018adversarially,pidhorskyi2018generative}, distance-based methods \citep{lee2018simple,ming2022cider,ren2021simple,sun2022out}, reconstruction-based methods \cite{li2023rethinking,denouden2018improving,zhou2022rethinking,yang2022out}, and gradient-based methods \citep{liang2017enhancing,huang2021importance,igoe2022useful}. Some works \citep{morteza2022provable, Fang2022IsOD, Bitterwolf2022BreakingDO} have also provided enhanced theoretical analyses for OOD detection. Recent studies have sought to reexamine and address the OOD detection problem from various perspectives. Ammar \textit{et al.} \citep{ammar2024neco} proposes utilizing "neural collapse" and the geometric properties of principal component spaces to identify OOD data. Lu \textit{et al.} \citep{lu2023learning} models each class with multiple prototypes to learn more compact sample embeddings, thereby enhancing OOD detection capabilities.

\paragraph{CLIP-based Out-of-distribution detection}
As multimodal models emerge as a new trend, multimodal pretrained models \citep{CLIP,ALIGN,groupvit} are increasingly being introduced into other domains \citep{yang2023diffusion,li2023blip,kirillov2023segment,radford2023robust,lin2023magic3d,fang2023eva}. The idea of utilizing large-scale pretrained multimodal models that rely solely on outlier class names without any accompanying images was first proposed in Fort \textit{et al.} \citep{fort2021exploring}. ZOC \citep{ZOC} employs an extended model to generate candidate unknown class names for each test sample. MCM\citep{MCM}, for the first time, abandons the traditional reliance on sample assumptions, starting from a language-vision representation and using a temperature scaling strategy and the maximum predicted softmax value as the out-of-distribution (OOD) score, establishing a paradigm for CLIP-based OOD detection. Building on MCM, GL-MCM \citep{glmcm} enhances the model by aligning the global and local visual-textual features of CLIP \citep{CLIP}.  LoCoOp\citep{LoCoOp} proposes a few-shot OOD detection method within the framework of prompt learning. LSN \citep{LSN}, NegPrompt \citep{NegPrompt}, and NegLabel \citep{NegLabel} all enhance the effectiveness of OOD detection by introducing negative prompts and utilizing methods such as prompt learning and sampling. Our approach, CMA, is rooted in MCM and offers a fresh perspective on CLIP-based OOD detection from the language-vision representation standpoint. It is a post-hoc detection method that obviates the need for learning and external data. 

\section{ID Classification Accuracy}
Table \ref{tab:acc} presents the multi-class classification accuracies achieved by various methods on ImageNet-1k. The findings indicate that CMA does not compromise the model's accuracy on the original classification task.

\section{Limitation}
Our proposed method, CMA, reconstructs the relationships in OOD detection by incorporating neutral prompts as agents, thereby enhancing the capability of OOD detection. However, the integration of external information is a double-edged sword. On one hand, if the neutral prompts share similar semantics with the ID labels, it can compromise the model's ID detection capabilities. On the other hand, external information struggles to mitigate the impact of OOD images that are too similar to the ID labels, as it does not directly augment the model's intrinsic abilities. Additionally, the challenge remains in obtaining neutral prompts that align more closely with the requirements of the ID labels without the need for training, necessitating further in-depth research.

\section{Experimental Details}
\label{sec:ap_exp_detail}

\subsection{Software and Hardware}
All methods are implemented in Pytorch 2.0.1 and Python 3.8. We run all OOD detection experiments on 4 NVIDIA RTX A6000 GPUs with 10 Intel Xeon Gold 5220 CPUs.

\subsection{Datasets}

\textbf{ImageNet-10} We use the ImageNet-10 in \citep{MCM} that mimics the class distribution of CIFAR-10 but with high-resolution images. It contains the following categories (with class ID): warplane (n04552348), sports car (n04285008), brambling bird, (n01530575), Siamese cat (n02123597), antelope (n02422699), Swiss mountain dog (n02107574), bull frog (n01641577), garbage truck (n03417042), horse (n02389026), container ship (n03095699).

\textbf{ImageNet-20} We use the ImageNet-20 in \citep{MCM}, which consists of 20 classes semantically similar  to ImageNet-10 (e.g. dog (ID) vs. wolf (OOD)). It contains the following categories: sailboat (n04147183), canoe (n02951358), balloon (n02782093), tank (n04389033), missile (n03773504), bullet train (n02917067), starfish (n02317335), spotted salamander (n01632458), common newt (n01630670), zebra (n01631663), frilled lizard (n02391049), green lizard (n01693334), African crocodile (n01697457), Arctic fox (n02120079), timber wolf (n02114367), brown bear (n02132136), moped (n03785016), steam locomotive (n04310018), space shuttle (n04266014), snowmobile (n04252077).

\textbf{ImageNet-100} We use the ImageNet-100 in \citep{MCM}, which is curated from ImageNet-1k. 

The aforementioned datasets can be accessed at \url{https://github.com/deeplearning-wisc/MCM}.

\subsection{Sources of model checkpoints}

For CLIP models, our reported results are based on checkpoints provided by Hugging Face for CLIP-B/16 \url{https://huggingface.co/openai/clip-vit-base-patch16} and CLIP-L/14 \url{https://huggingface.co/openai/clip-vit-large-patch14}. Similar results can be obtained with checkpoints in the codebase by OpenAI \url{https://github.com/openai/CLIP}. Note that for CLIP-RN50, which is not available in Hugging Face, we use the checkpoint provided by OpenAI. 

\subsection{Hyperparameters}

In our approach, in addition to the number of agents, $M$, which we have already discussed in the Discussion section (See Section \ref{sec:discussion}), an important hyperparameter is temperature parameter $\tau$. In all the experiments presented in the main text, $\tau$ is set to 1. This is because our empirical findings (Table \ref{table:t}) indicate that CMA is insensitive to $\tau$, resulting in minimal variations in outcomes.

\begin{table*}[t]
\centering
\footnotesize
\caption{Comparison results of CMA with different Temperature parameter $\tau$. The ID dataset is ImageNet-1k.}
\label{table:t}
\footnotesize
\centering
     {\resizebox{\textwidth}{!}{
    \begin{tabular}{@{}lcccccccccccccc@{}} \toprule
    & \multicolumn{2}{c}{iNaturalist} && \multicolumn{2}{c}{SUN} && \multicolumn{2}{c}{Places} && \multicolumn{2}{c}{Texture} && \multicolumn{2}{c}{\textbf{Average}} \\ 
    \cmidrule(lr){2-3} \cmidrule(lr){5-6} \cmidrule(lr){8-9} \cmidrule(lr){11-12} \cmidrule(lr){14-15}
   \textbf{$\tau$} & \scriptsize FPR95$\downarrow$ & \scriptsize AUROC$\uparrow$ && \scriptsize FPR95$\downarrow$ & \scriptsize AUROC$\uparrow$ && \scriptsize FPR95$\downarrow$ & \scriptsize AUROC$\uparrow$ && \scriptsize FPR95$\downarrow$ & \scriptsize AUROC$\uparrow$ && \scriptsize FPR95$\downarrow$ & \scriptsize AUROC$\uparrow$  \\ 
   
    \midrule
    0.1 & 21.20 & 95.94 && 32.21 & 93.08 && 30.33 & 93.20 && 49.04 & 88.11 && 33.00 & 92.58 \\
    0.2 & 22.42 & 95.70 && 31.10 & 93.26 && 29.95 & 93.20 && 47.89 & 88.32 && 32.49 & 92.62 \\
    0.4 & 23.22 & 95.55 && 30.73 & 93.34 && 29.88 & 93.19 && 29.88 & 88.42 && 28.89 & 92.63 \\
    0.6 & 23.64 & 95.49 && 30.64 & 93.37 && 29.88 & 93.18 && 47.48 & 88.45 && 32.46 & 92.62 \\
    0.8 & 23.75 & 95.46 && 30.55 & 93.38 && 29.85 & 93.18 && 47.39 & 88.47 && 32.42 & 92.62 \\
    1 & 23.84 & 95.44 && 30.51 & 93.39 && 29.86 & 93.17 && 47.35 & 88.47 && 32.41 & 92.62 \\
    1.5 & 23.99 & 95.42 && 30.49 & 93.40 && 29.88 & 93.17 && 47.32 & 88.49 && 32.43 & 92.62 \\
    2 & 24.02 & 95.41 && 30.47 & 93.40 && 29.84 & 93.17 && 47.26 & 88.49 && 32.41 & 92.62 \\
    4 & 24.14 & 95.39 && 30.39 & 93.41 && 29.85 & 93.16 && 47.25 & 88.50 && 32.40 & 92.62 \\
    8 & 24.21 & 95.38 && 30.43 & 93.41 && 29.89 & 93.16 && 47.23 & 88.51 && 32.44 & 92.62 \\
    16 & 24.23 & 95.37 && 30.44 & 93.41 && 29.90 & 93.16 && 47.23 & 88.51 && 32.45 & 92.61 \\
    32 & 24.29 & 95.37 && 30.45 & 93.42 && 29.94 & 93.16 && 47.26 & 88.51 && 32.48 & 92.62 \\
    64 & 24.27 & 95.37 && 30.42 & 93.42 && 29.94 & 93.16 && 47.25 & 88.51 && 32.46 & 92.62 \\
    \bottomrule
    \end{tabular}
    }
    }

\end{table*}

\begin{table*}[t]
\caption{Comparison results of ID classification accuracy on ImageNet-1k (\%)}
\label{tab:acc}
\centering
\small
\begin{tabular}{lc}
\toprule
\textbf{Method} & \textbf{ID ACC}\\
\midrule
\multicolumn{2}{c}{\textbf{zero-shot }}   \\ 
MSP (CLIP-B/16) \citep{hendrycks2016baseline} & 66.78 \\
MSP (CLIP-L/14) \citep{hendrycks2016baseline} & 73.12 \\
MCM (CLIP-B/16) \citep{MCM} &67.01 \\
MCM (CLIP-L/14) \citep{MCM} & 73.28\\
CMA (CLIP-B/16) &66.88 \\
CMA (CLIP-L/14) & 73.49\\
\bottomrule
\end{tabular}
\end{table*}

\begin{figure*}
\lstset{
framesep=20pt,
rulesep=10pt,
backgroundcolor=\color[RGB]{255,255,255},
breaklines=true,
numbers=none,
breakindent=0pt,
basicstyle=\ttfamily\small,
escapeinside={(*@}{@*)}, 
}

\subsection{Prompt-1 to generate Neutral Prompts}
\begin{lstlisting}
Create a series of neutral, generic image prompts that do not specify any particular object, scene, or concept. These prompts should be vague and open-ended, allowing for a wide range of possible interpretations. Follow these guidelines:

1. Start each prompt with "a photo of" or "an image of".
2. Use generic terms like "thing", "object", "item", or "element".
3. Optionally, add a broad context like "in nature", "in a place", or "in an environment".
4. Avoid mentioning any specific objects, living beings, or locations.
5. Keep the prompts simple and concise, typically 5-10 words long.
6. Do not include any adjectives that could narrow down the interpretation.
7. Ensure the prompts are grammatically correct and make sense in English.
8. Create at least 10 unique prompts following these rules.

Examples of the type of prompts you should generate:
- "a photo of a thing in the world"
- "an image of an object somewhere"
- "a picture of something that exists"
- "a photograph of an item in a setting"

Remember, the goal is to create prompts that are as neutral and non-specific as possible while still forming a coherent phrase about an image or photo.
\end{lstlisting} \label{Prompt-1}
\end{figure*}

\begin{figure*}
\lstset{
framesep=20pt,
rulesep=10pt,
backgroundcolor=\color[RGB]{255,255,255},
breaklines=true,
numbers=none,
breakindent=0pt,
basicstyle=\ttfamily\small,
escapeinside={(*@}{@*)}, 
}

\subsection{Prompt-2 to generate Neutral Prompts}
\begin{lstlisting}
Create a series of neutral, abstract image prompts that do not specify any particular object, scene, or concept. These prompts should be vague, open-ended, and focus on general concepts or sensory experiences. Follow these guidelines:

1. Start each prompt with "a photo of", "an image of", or a similar phrase.
2. Use generic terms like "thing", "object", "item", "element", or more abstract concepts like "presence" or "essence".
3. Optionally, add a broad context like "in nature", "in a place", or "in an environment".
4. Avoid mentioning any specific objects, living beings, or locations.
5. Keep the prompts simple and concise, typically 5-15 words long.
6. Do not include any adjectives that could narrow down the interpretation.
7. Ensure the prompts are grammatically correct and make sense in English.
8. Focus on general concepts rather than specific items.
9. Use neutral descriptive language that evokes feelings or imagery without defining what is being referred to.
10. Include references to sensory experiences (sight, sound, touch, smell, taste) without attributing them to particular phenomena.
11. Use broad terms instead of elements that could be interpreted as belonging to a particular category.
12. Employ indefinite articles (a/an) and vague quantifiers (some, many) to maintain ambiguity.
13. Create at least 10 unique prompts following these rules.

Examples of the type of prompts you should generate:
- "a photo of a thing in the world"
- "an image of an abstract presence in a space"
- "a picture of something evoking a gentle sensation"
- "a photograph of an undefined texture in an environment"
- "an image capturing a vague impression of color"

Remember, the goal is to create prompts that are as neutral, abstract, and non-specific as possible while still forming a coherent phrase about an image or photo. Focus on sensory elements and general concepts without implying any specific meaning or object.
\end{lstlisting} \label{Prompt-2}
\end{figure*}

\end{document}